\pdfoutput=1

\documentclass[11pt]{article}

\usepackage[preprint]{acl}

\definecolor{nicolo}{RGB}{150, 0, 0}
\definecolor{nicolo2}{RGB}{150, 0, 150}
\definecolor{maryam}{RGB}{0, 150, 0}
\definecolor{sara}{RGB}{0, 0, 150}
\definecolor{marco}{RGB}{150, 150, 0}
\definecolor{bruno}{RGB}{0, 150, 150}

\usepackage{times}
\usepackage{latexsym}
\usepackage{verbatim}
\usepackage{amsmath}
\usepackage{booktabs}
\usepackage{hyperref}
\usepackage{todonotes}
\usepackage{tabularx}
\usepackage{amssymb,graphicx}
\usepackage[T1]{fontenc}

\usepackage[utf8]{inputenc}

\usepackage{microtype}

\usepackage{inconsolata}

\usepackage{graphicx}

\title{\textit{Do LLMs suffer from Multi-Party Hangover?} \\A Diagnostic Approach to Addressee Recognition and Response Selection\\ in Conversations}

\author{Nicol\`o Penzo$^{1, 2}$, Maryam Sajedinia$^{1, 3}$, Bruno Lepri$^1$, Sara Tonelli$^1$, Marco Guerini$^1$\\
  $^1$ Fondazione Bruno Kessler, Italy\\
  $^2$ University of Trento, Italy \\
  $^3$ University of Turin, Italy \\
  \texttt{\{npenzo,msajedinia,lepri,satonelli,guerini\}@fbk.eu}
  }

\begin{document}
\maketitle
\begin{abstract}

Assessing the performance of systems to classify Multi-Party Conversations (MPC) is challenging due to the interconnection between  linguistic and structural characteristics of conversations. Conventional evaluation methods often overlook variances in model behavior across different levels of structural complexity on interaction graphs.
In this work, we propose a methodological pipeline to investigate model performance across specific structural attributes of conversations. As a proof of concept we focus on Response Selection and Addressee Recognition tasks, to diagnose model weaknesses. 
To this end,  we extract representative diagnostic subdatasets with a fixed number of users and a good structural variety from a large and open corpus of online MPCs. 
We further frame our work in terms of data minimization, avoiding the use of original usernames to preserve privacy, and propose alternatives to using original text messages.
Results show that response selection relies more on the textual content of conversations, while addressee recognition requires capturing their  structural dimension.
Using an LLM in a zero-shot setting, we further highlight how sensitivity to prompt variations is task-dependent.

\end{abstract}

\section{Introduction}
\label{sec:intro}

Multi-Party Conversations (MPCs) are multi-turn discussions involving more than two participants \cite{traum2003issues, branigan2006perspectives}, which are typical of online platforms such as Reddit or Twitter/X  \cite{mahajan2021need}. Being able to capture the content of such discussions, when multiple users are involved and the conversation is composed of several turns, is a challenging task because both textual and structural information need to be modeled. Indeed, designing systems for MPC understanding is challenging not only because the \textit{textual} dimension spans multiple turns, but also because we need to capture \textit{structural} aspects, such as who writes to whom. Understanding how these two  components should be integrated to classify MPC, and how effectively LLMs can contribute to this task, is still an open question.

\begin{figure}
    \centering
    \includegraphics[width=0.43\textwidth]{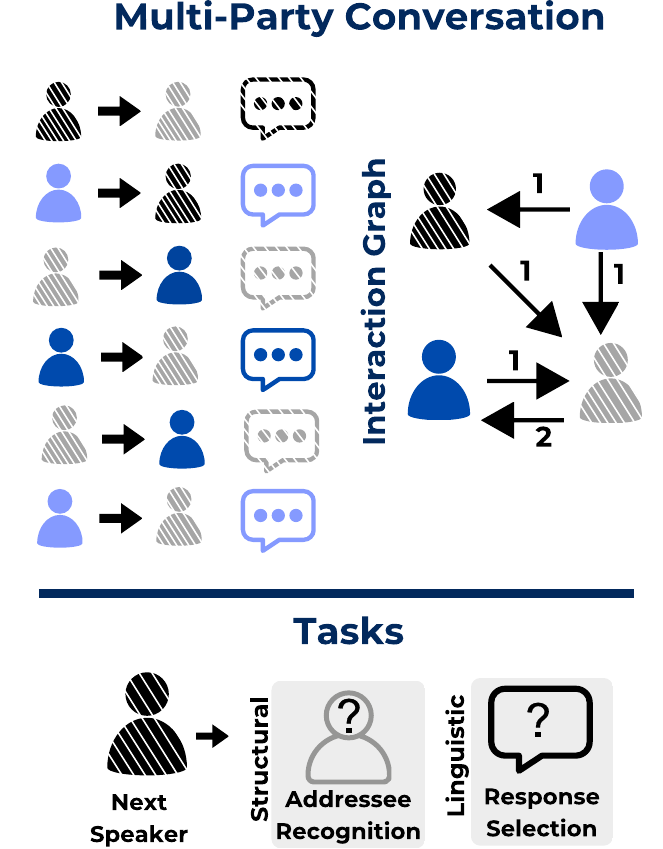}
    \caption{A graphical representation of the experiments. Each turn in a conversation includes a speaker, an addressee and a textual message. From the conversation, we extract the interaction graph to diagnose model capabilities by performing two tasks: addressee recognition and response selection. 
    }
    \label{fig:first_image}
\end{figure}

In this paper, we examine the ability of an LLM to perform MPC classification tasks in a zero-shot setting as well as to capture relevant information from an existing conversation. Specifically, we address the tasks of \textit{Response Selection} and \textit{Addressee Recognition} and we use \texttt{Llama2-13b-chat} \cite{touvron2023llama} not only to classify the last turn of MPCs but also to summarise the previous conversation and to describe users, so that this information can be included in the prompts for zero-shot classification. Our choice of these two classification tasks is based on the following key considerations: \textsc{i.} the tasks deal with two specific aspects that a model working on MPCs needs to address, i.e. response selection for linguistic aspects and addressee recognition for structural and non-linguistic aspects; \textsc{ii.} these tasks can be performed on any conversational corpus in any domain, without the need of manual annotation. This makes our framework widely applicable. In Figure \ref{fig:first_image} we report a graphical representation of an MPC, the pieces of information we retrieve and the tasks we perform.

Understanding the effects of  conversation summarisation and user descriptions is important because they could make processing more efficient, replacing multiple turns with a more concise text representation. Furthermore, using summarisations and user descriptions instead of the original conversations would make data sharing easier and more privacy-preserving, addressing growing concerns about this issue \cite{NEURIPS2023_420678bb}. For instance, it would comply with data minimisation principles, as required by the European General Data Protection Regulation. 
Replacing original conversations with summaries and user descriptions would also make it nearly impossible to train generative models that imitate specific users \cite{huang-etal-2022-mcp, lu-etal-2023-miracle}. 

Our research questions are therefore as follows:

\begin{itemize}
\item[] \textbf{RQ(1)}: How do LLMs perform in classification tasks involving MPCs in a zero-shot setting, using different input combinations to capture textual and structural information?
\item[] \textbf{RQ(2)}: What is the model  sensitivity to different prompt formulations when classifying MPCs?
\item[] \textbf{RQ(3)}: How does structural complexity of the conversation affect classification performance?
\end{itemize}

To address \textbf{RQ(1)}, we evaluate \texttt{Llama2-13b-chat} on response selection and addressee recognition in a zero-shot scenario (Section \ref{sec:tasks}). These tasks capture different types of information: response selection relies on textual information to choose the next message in a conversation, while addressee recognition requires more structural awareness to infer speaker characteristics and conversation flow. 
For each conversation, we design input combinations of conversation transcripts, interaction transcripts, generated summaries, and generated user descriptions, with the latter two being generated by \texttt{Llama2-13b-chat} (Section \ref{sec:workflow}). We address also \textbf{RQ(2)} by designing prompts of different levels of verbosity for each combination and task.
Finally \textbf{RQ(3)} is addressed by designing a diagnostic approach, where the two tasks are evaluated on MPCs with a different number of speakers and structural characteristics (Section \ref{sec:diagnostic_pipeline}). This allows us to analyse the connection between task scores and structural characteristics of MPCs. 

The software to perform the experiments and the processed data are available on a dedicate Github repository.\footnote{\url{https://github.com/dhfbk/MPH}}

\section{Related Work}

Researchers have worked on MPC understanding tasks either by trying to model an entire conversation or by focusing on relations within the conversation \cite{gu2022says, ganesh-etal-2023-survey}. Recent MPC understanding studies focus on response selection (RS) and addressee recognition (AR) tasks \cite{ouchi-tsuboi-2016-addressee, Zhang_Lee_Polymenakos_Radev_2018} to compare different classification approaches. Indeed, RS is strictly related to textual (linguistic) information, while AR focuses on interaction information, thus permitting to analyse the performance of classification models from two different angles. However, both tasks can ideally benefit from cross-information between linguistic and interaction cues.

For both RS and AR, researchers have fine-tuned transformer-based models incorporating speaker information \cite{wang-etal-2020-response, gu-etal-2021-mpc, zhu-etal-2023-robust, gu-etal-2023-gift}, used Graph Neural Networks (GNNs) for interaction modeling \cite{ijcai2019p696, gu-etal-2022-hetermpc}, or leveraged dialogue dependency parsing \cite{jia-etal-2020-multi}. Recently, \citet{tan-etal-2023-chatgpt}, explored zero-shot capabilities of ChatGPT \cite{openai_chatgpt} and GPT-4 \cite{openai2024gpt4} in MPCs,  focusing only on the overall classification performance. Indeed, there is a gap in the NLP literature concerning the evaluation of MPC systems based on structural aspects. Past research has focused on textual information, for instance by using candidate rankings \cite{mahajan-etal-2022-towards} or just looking at conversation length and number of users \cite{gu-etal-2023-gift}. \citet{penzo-etal-2024-putting} provided a first exploration of the role of conversation structure in stance detection, showing that it benefits classification only when large training data are available.

Summarizing the dynamics and trajectories of MPCs, where a model's understanding of the conversation structure and interactions is critical, has been recently addressed by \citet{hua-etal-2024-get}. The authors also evaluate the summaries of conversation dynamics with a classification task (i.e., forecasting the future derailment of the conversation as in \citealp{zhang-etal-2018-conversations}). \citet{hua-etal-2024-get} point out that conventional summaries heavily focus on the textual content and what individual speakers say while ignoring the interactions between speakers and the conversation flow.

The work that is most similar to our contribution is \citet{tan-etal-2023-chatgpt}, since they also use a generative model in a zero-shot setting to address RS and AR.
The main difference is that, instead of focusing only on generic accuracy scores, we propose a diagnostic approach for evaluating models for MPC understanding. We use response selection and addressee recognition as proxy tasks and focus particularly on the contribution of structural information, by \textsc{i.} creating diagnostic datasets, each with a fixed number of users, and \textsc{ii.} putting in relation the classification performance to specific network metrics (i.e., degree centrality and average outgoing weight of the speaker node).

\section{Tasks}\label{sec:tasks}

Our experiments revolve around two tasks that do not need a manual annotation as long as the used MPC data include speaker, addressee and related utterances.

\textbf{Response Selection} 
(RS) is the task of choosing the text of the next message given a conversation $C$, the id of the speaker of the next message and a set of candidate responses. In our experiments, we cast response selection as a binary classification task, since the system has to choose between two possible candidates (similar to the R2@1 task in \citealp{gu-etal-2021-mpc}).  

\textbf{Addressee Recognition} 
(AR) is the task of predicting the addressee of the next message given a conversation $C$, the id of the speaker of the next message and a set of candidate addressees. The set of candidate addressees include all speakers involved so far in the conversation $C$ plus a ``dummy'' option, which introduces a user unseen in the conversation to check whether the classifier choice is fully random.

\vspace{0.1cm}
In both cases, the next speaker is given, and the classifier has to select what will be the content of the message (RS) or who will be the addressee (AR).

\section{MPC Classification Workflow}\label{sec:workflow}

In this section we describe the classification workflow implemented to perform response selection (RS) and addressee recognition (AR). The workflow is shared between the two tasks.

\begin{figure}
    \centering
    \includegraphics[width=\linewidth]{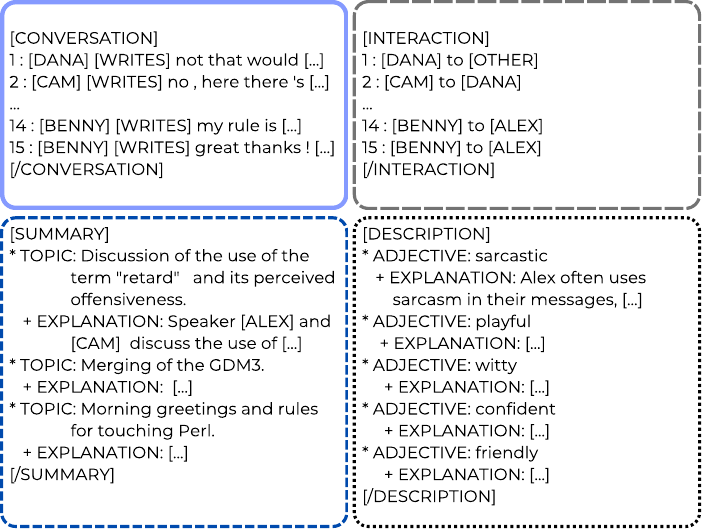}
    \caption{Example of the $4$ possible conversation representations: \textsc{i.} Conversation Transcript (top left), \textsc{ii.} Interaction Transcript (top right), \textsc{iii.} Summary (bottom left) and \textsc{iv.} User Description (bottom right).}
    \label{fig:conv_repr}
\end{figure}

\subsection{Conversation Representation}
\label{convrep}

The first step is modelling the input data 
to be included in the prompt used for classification. 
To analyse the contribution of contextual and structural information for RS and AR 
we identify four ways to model the conversation content. 
The first includes just the chronologically ordered list of speaker-message pairs. This input format is called \textbf{(i) Conversation Transcript}.

The second way aims at including only structural information to assess its contribution in the classification tasks when no textual content is given. We call it \textbf{(ii) Interaction Transcript}, and we model it as a chronologically ordered list of speaker-addressee pairs without the actual turn content.

The third and fourth settings aim at assessing how reliable LLMs are in representing a sequence of turns and capturing the most relevant information. We prompt an LLM to provide two types of output, given the Conversation Transcript and the Interaction Transcript: \textbf{(iii) Summary} of the conversation, expressed by the three main topics discussed, each followed by a brief explanation, and \textbf{(iv) User Description}, i.e., a description of the behavior of the next speaker inside the given conversation, using five adjectives with a brief explanation for each. An example of each type of conversation representation is reported in Figure \ref{fig:conv_repr}.

These last two representations are meant to replace the actual discussion content, retaining only the most relevant information. This approach can be useful in settings where storing and/or classifying whole conversations may be too expensive or when the actual conversation may become unavailable or impossible to reshare. Distributing raw conversational data with user IDs and full messages could in fact lead to potential malicious use, such as user profiling \cite{wen-etal-2023-towards} or training LLMs to create fake personas \cite{huang-etal-2022-mcp}. To ensure anonymization and avoid gender bias in classifier decisions, the original conversations are pre-processed by replacing real usernames with fake gender-neutral names, forcing the model to perform the MPC tasks using only ``local'' users \cite{penzo-etal-2024-putting}, so that it is not possible to identify which users are the same across different conversations (details in Appendix \ref{sec:app_prompt_design}).

\subsection{Pipeline and Prompt Design}

We use \texttt{Llama2-13b-chat} \cite{touvron2023llama} to perform text generation. Specifically, it is employed in four steps of our workflow: \textsc{i.} to generate a summary of each conversation; \textsc{ii.} to generate user descriptions for each conversation; and then for zero-shot classification, namely \textsc{iii.} response selection and \textsc{iv.} addressee recognition. 
For creating prompts, we follow the guidelines provided by Meta\footnote{\url{https://llama.meta.com/get-started/\#prompting}}. 

In \texttt{Llama2-13b-chat}, each prompt is composed by a system prompt $s$ that describes the task concatenated to an input prompt $i$ that provides input information and the instruction command (i.e., the command to start the task to perform). For performing the generation of summaries and user descriptions, we use a greedy decoding mechanism and we design a generation prompt $p_g$ with the following structure:

\verb|[INST] <<SYS>> s <</SYS>> i [/INST]|

Instead, for the two classification tasks, the candidate responses are given. So, instead of having the LLM generate the output response, we evaluate the Conditional Perplexity, CPPL \cite{su-etal-2021-put, occhipinti2023prodigy} of all candidate responses given the classification prompt $p_c$, selecting as best output the candidate with the lowest CPPL. Other works dealing with classification tasks compute the probability of each candidate instead of CPPL \cite{liusie-etal-2024-llm}. However, probability can be applied to settings where each candidate includes only one word, whereas in our response selection task the candidates are sentences of variable length.

Each classification prompt $p_c$ includes a system prompt $s$ and an input prompt $i$. Moreover, we add a ``beginning of output'' prompt $b$, in order to evaluate CPPL only on the candidate responses.
The prompt $p_c$ presents the following structure :

\vspace{0.2cm}
\verb|[INST]<<SYS>>| $s$ \verb|<</SYS>>| $i$ \verb|[/INST]| $b$

\vspace{0.2cm}
which leads to the full prompt $p_{c_i}$ with the candidate responses $r_i$ being:

\vspace{0.2cm}
\verb|[INST]<<SYS>>| $s$ \verb|<</SYS>>| $i$ \verb|[/INST]| $b$ $r_i$ 

\subsection{Prompt Details}\label{subsec:prompt_design}

\begin{figure}
    \centering
    \includegraphics[width=0.95\linewidth]{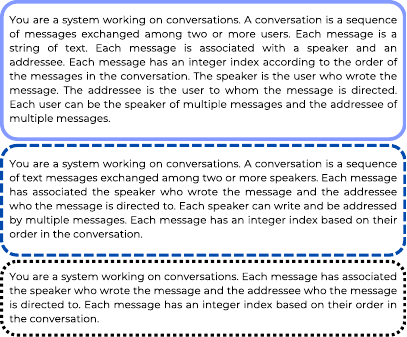}
    \caption{Example of the beginning of the system prompt in the three prompt schemes, from the most verbose (top) to the most concise (bottom).}
    \label{tab:example_beginning}
\end{figure}

We compare three distinct prompt schemes with varying levels of verbosity to test LLM classification robustness and prompt sensitivity \cite{sun2024evaluating}. Each prompt varies in terms of being more or less explicit in providing information. This leads us to have, for each prompt for a specific input combination and task, one \textit{verbose} version, one \textit{concise} version and one \textit{medium} version. Our hypothesis is that the verbose prompts, giving more detailed instructions 
can potentially improve classification performance of \texttt{Llama2-13b-chat}. Figure \ref{tab:example_beginning} shows how the beginning of the system prompt changes across the three versions. More details and examples are provided in Appendix \ref{sec:app_prompt_design}.

\section{Diagnostic Approach}\label{sec:diagnostic_pipeline}

\begin{figure*}
    \centering
    \includegraphics[width=0.66\linewidth]{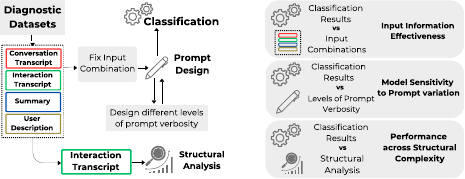}
    \caption{Schematic representation of our evaluation pipeline: on the left, the pipeline and the relation among the elements; on the right, the type of diagnostic evaluation we can perform.}
    \label{fig:pipeline}
\end{figure*}

To address the three research questions introduced in Section \ref{sec:intro}, we aim to develop a diagnostic approach that isolates specific phenomena and minimizes confounding factors. A key aspect under analysis is the interplay between interaction structure in the conversation and classification performance. We identify two metrics to capture conversation complexity in terms of interaction graph and we also create subcorpora from a large conversation corpus, called \textit{diagnostic datasets}, each with specific characteristics to test in relation to classification performance. In Figure \ref{fig:pipeline} we report a schematic representation of our evaluation pipeline and the components involved.

\subsection{Structural Information as Interaction Graph}
\label{sec:struc}

Conversations present different structures depending on the discussion complexity and the speakers' involvement \cite{10.1145/2392622.2392626, COLETTO201722}. To analyse the relation between interaction complexity and classification  performance, we first identify two network metrics able to capture the structural complexity of MPCs.
In the past, researchers have explored correlations between model performance and factors such as number of speakers and conversation length \cite{gu-etal-2023-gift, penzo-etal-2024-putting}, but to our knowledge network metrics to capture conversation complexity have never been considered before in this framework.  

Given an MPC, its interaction graph can be modeled as an  \textsc{i.} \textit{unweighted undirected}; or \textsc{ii.} \textit{weighted directed} graph.
In the unweighted undirected graph, each edge between two users simply indicates that they have interacted, without specifying the direction of the communication and the number of exchanged messages.
The weighted directed graph instead includes directionality from the speaker to the addressee and a weight assigned to each edge, corresponding to the number of messages from the speaker to the addressee. 
From each conversation $C$ we therefore extract both versions of the interaction graph, i.e. $G_{ud}^{uw}(C)$ and $G_{d}^{w}(C)$.

From the above interaction graph, we then derive two network metrics for the next speaker node, i.e. the degree centrality and the average weight of the outgoing edges (average outgoing weight). 
Specifically, the metrics are computed as follows:

\textbf{Degree Centrality.} Given an unweighted undirected graph $G_{ud}^{uw}(C) = (U, E)$, where $U$ is the set of nodes and $E$ is the set of edges, the degree centrality $deg(u)$ of a node $u \in U$ is the number of edges $e \in E$ incident in $u$, i.e. $e = (u, v)$ or $e = (v, u)$. In our setting, it represents the number of users the next speaker has interacted with.

\textbf{Average Outgoing Weight.} Given a weighted directed graph $G_{d}^{w}(C) = (U, E)$, where $U$ is the set of nodes and $E$ is the set of edges, the out-degree centrality $outdeg(u)$ of a node $u \in U$ is the number of outgoing edges $e \in E$ for which $u$ is the originating node/speaker, i.e. $e = (u, v)$ and the weighted out-degree $outdeg_w(u)$ is the sum of the weights on such edges. So the average outgoing weight $w_{avg}^o(u) = outdeg_w(u)/outdeg(u)$ is the average weight of the edges $e \in E$ for which $u$ is the originating node/speaker, i.e. $e = (u, v)$. In our setting, it represents the average number of messages sent to the users with whom the next speaker has interacted.

\subsection{Diagnostic Datasets}
To analyse the impact that different interaction structures have on RS and AR, we create four datasets derived from the Ubuntu Internet Relay Chat corpus \cite{ouchi-tsuboi-2016-addressee}, which includes more than 800,000 conversations in English about how to solve technical issues. 
We used such large corpus because, to the best of our knowledge, it is the only one with an adequate dimension to allow us to extract a good number of diagnostic MPCs with: \textsc{i.} a defined number of users; \textsc{ii.} a good length of discussion; \textsc{iii.} a good structural variety, for each "diagnostic" subsets. Moreover, it involves natural conversations with explicit addressee, which are necessary for the AR task.

To control the fluctuations in  structural complexity, we limit the maximum conversation length to $15$ messages (in line with the Len-15 version in \citealp{gu-etal-2021-mpc}). We then create $4$ MPC diagnostic subsets with conversations involving $3$, $4$, $5$ and $6$ users, which we call respectively \textit{Ubuntu3/4/5/6}. 
Then, for all $4$ subsets, we proceed as follows: \textsc{i.} for each conversation, we extract the undirected and unweighted interaction graph as explained above; \textsc{ii.} we keep only the conversations where the corresponding undirected and unweighted interaction graph is connected.

Finally, we anonymize the users, by replacing each username with a fake username, as already mentioned in Section \ref{convrep} (details in Appendix \ref{sec:app_prompt_design}).  
 
The resulting diagnostic datasets have respectively $1200$, $635$, $520$ and $350$ conversations. 
These datasets are used as test sets for evaluating RS and AR in a zero-shot setting. 

\section{Experiments}

Given the four types of input presented in Section \ref{convrep}, we design five input combinations to test in our prompts for both tasks: \textsc{i.} only the conversation transcript (\textsc{CONV}); \textsc{ii.} the conversation transcript and the interaction transcript (\textsc{CONV+STRUCT}); \textsc{iii.} the interaction transcript 
 and the conversation summary (\textsc{STRUCT+SUMM}); \textsc{iv.} the interaction transcript 
 and the user descriptions (\textsc{STRUCT+DESC}); \textsc{v.} the interaction transcript, the conversation summary  
 and the user descriptions (\textsc{STRUCT+SUMM+DESC}). 
For the AR task, we test a sixth combination, \textsc{vi.} \textsc{STRUCT}, which corresponds only to the interaction transcript. \textsc{STRUCT} is not relevant for RS since it does not include any linguistic information.

All combinations and prompt schemes are tested across the  $4$ diagnostic datasets. 

\begin{figure*}[ht]
    \centering
    \includegraphics[width=0.95\textwidth]{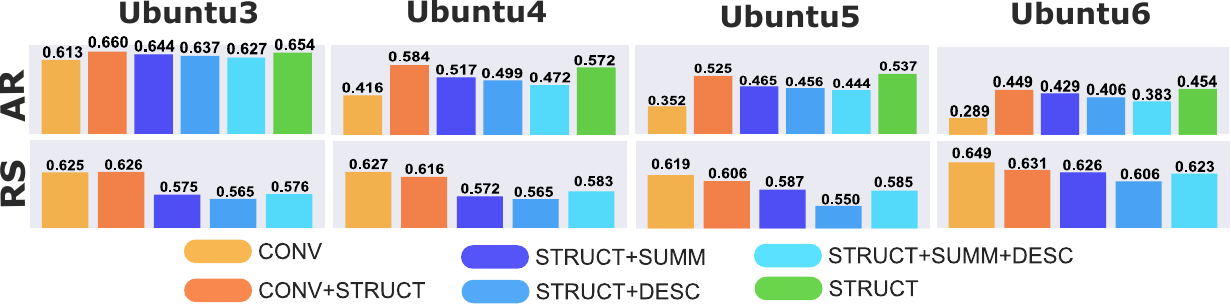}
    \caption{AR and RS macro-accuracy results ($y$ axis), for each combination and for each dataset. The height of the columns represents the best macro result across the three prompt schemes. Note that for AR the number of classes on each Ubuntu subset changes, ranging from four (Ubuntu3) to seven (Ubuntu6), since the set of possible addressees includes the speakers involved in each conversation, plus the dummy label. For this reason, results across different Ubuntu subsets on AR should not be compared, and the lowest accuracy is achieved on Ubuntu6.}
    \label{fig:macro_results}
\end{figure*}

\section{Macro Results and Structural Evaluation}\label{sec:results}

\vspace{0.2cm}
\textbf{Macro-results on the best run.} 

In Figure \ref{fig:macro_results}, we present the macro accuracy for both tasks across all $4$ diagnostic datasets. The columns show the highest accuracy achieved among the $3$ prompt schemes with varying level of verbosity.

In the \textbf{AR task},  the number of classes (i.e. addressees) on each Ubuntu subset changes, ranging from four (Ubuntu3) to seven (Ubuntu6), since the set of possible addressees includes the speakers involved in each conversation, plus the dummy label (see Section \ref{sec:tasks}). For this reason, results across different Ubuntu subsets on AR should not be compared, and the lowest accuracy is achieved on Ubuntu 6, being its classification based on seven possible addressees.

In AR, the \textsc{CONV+STRUCT} and \textsc{STRUCT} combinations consistently perform best across all datasets. 
Instead, the \textsc{CONV} combination, serving as our `text-only' baseline consistently shows the worst performance. 
If we consider replacing the original conversation with a summary (SUMM) or user description (DESC), we observe that the former outperforms the other on all datasets, although adding the original conversation to the structure (CONV+STRUCT) still ourperforms both alternatives.

In the \textbf{RS task}, the \textsc{CONV} and \textsc{CONV+STRUCT} combinations consistently perform the best across all datasets. 
Among the combinations with summary and/or user description, \textsc{STRUCT+SUMM+DESC} performs the best (in Ubuntu3/4) or extremely close to \textsc{STRUCT+SUMM}, which is the best in Ubuntu5/6. 
Finally, the \textsc{STRUCT+DESC} input combination yields the lowest classification performance for both tasks on all datasets, except for AR on Ubuntu3. 

This analysis shows that the interaction transcript (i.e., the structural information) is fundamental for achieving the best result in AR. On the other hand, the conversation transcript is fundamental for achieving the best results in RS, which in fact is a more text-oriented task, based on information mainly available in the conversation itself. However, using summaries of conversation may be a viable alternative, achieving results closer to the best input combinations in the setting with more users (i.e. Ubuntu6) for both tasks.

\vspace{0.2cm}
\textbf{Prompt Sensitivity.} 
In this section we compare the highest accuracy and average accuracy among the $3$ prompt schemes, for each input combination. Given $b$ the accuracy of the best prompt scheme and $a$ the average among the $3$ prompt schemes, we define as \textit{relative gap} the relative worsening from the best to the average: $gap_{rel}=1-a/b$. A larger relative gap suggests greater sensitivity of the model to the prompts used, which leads to fluctuations in the classification results.

\begin{table}[ht]
\small
\centering
\begin{tabular}{|l|c|l|l|l|l|}
\hline
\textbf{COMBINAT.} & \textbf{T} & \textbf{U.3} & \textbf{U.4} & \textbf{U.5} & \textbf{U.6} \\
\hline
\textsc{CONV} & AR & $2.7^{\diamond}$ & $0.6$ & $5.8^{\diamond}$ & $2.0$ \\
     & RS & $0.8^{\diamond}$ & $0.9^{\diamond}$ & $0.8^{\diamond}$ & $0.6$ \\
\hline
\textsc{CONV +} & AR & $7.1^{\diamond}$ & $10.9^{\diamond}$ & $4.5^{\diamond}$ & $4.9^{\diamond}$ \\
        {STRUCT}    & RS & $0.4^{\diamond}$ & $0.6$ & $1.1^{\ast}$ & $0.8^{\ast}$ \\
\hline
\textsc{STRUCT+} & AR & $2.5^{\ast}$ & $6.5^{\diamond}$ & $3.6^{\ast}$ & $6.7^{\ast}$ \\
\textsc{SUMM}   & RS & $0.8^{\ast}$ & $1.0$ & $1.7$ & $0.8^{\diamond}$ \\
\hline
\textsc{STRUCT+} & AR & $2.7^{\diamond}$ & $5.6^{\diamond}$ & $4.8^{\diamond}$ & $8.9^{\diamond}$ \\
  \textsc{DESC}  & RS & $2.1^{\diamond}$ & $0.9^{\ast}$ & $1.3^{\ast}$ & $1.6$ \\
\hline
\textsc{STRUCT+} & AR & $1.5^{\diamond}$ & $4.0^{\diamond}$ & $3.2^{\ast}$ & $6.0^{\diamond}$ \\
 \textsc{SUMM+DESC} & RS & $0.2^{\diamond}$ & $1.4$ & $1.2$ & $0.6^{\diamond}$ \\
\hline
\textsc{STRUCT} & AR & $5.6^{\diamond}$ & $8.3^{\diamond}$ & $8.5^{\diamond}$ & $6.3^{\diamond}$ \\
\hline
\end{tabular}
\caption{Relative gap (\%) between the best prompt result and the average, for each input combination and diagnostic dataset (UbuntuX is shortened as U.X), and for each task (i.e., AR and RS). We put a $\diamond$ when the best prompt is the verbose version, a $\ast$ when the medium version is the best and nothing when the best is the concise version. 
}
\label{tab:prompt_gap}
\end{table}

In Table \ref{tab:prompt_gap} we report the relative gaps between accuracy achieved with the best prompt and the average accuracy obtained for each input combination and diagnostic dataset. Results on the AR task tend to be more sensitive to prompt formulation compared to the RS task, especially in the \textsc{CONV+STRUCT} combination. Indeed, the relative gap between the best run and the average results is remarkably larger for the AR task than for RS across all diagnostic datasets. 
Moreover, in the AR task, the \textsc{STRUCT} combination has similar prompt sensitivity to \textsc{CONV+STRUCT}. 

Overall, we observe that for AR, where structural information is more relevant, 
classification performance tends to vary more with different prompt verbosity compared to RS, where linguistic information has a higher weight. 

If we analyse what is the effect of prompt verbosity on classification performance, we observe that in the majority of settings and configurations, the verbose version of the prompt is the best performing one for AR. For RS, instead, there is no evidence of benefit from using more or less verbose options.

\begin{figure*}
    \centering
    \includegraphics[width=0.99\textwidth]{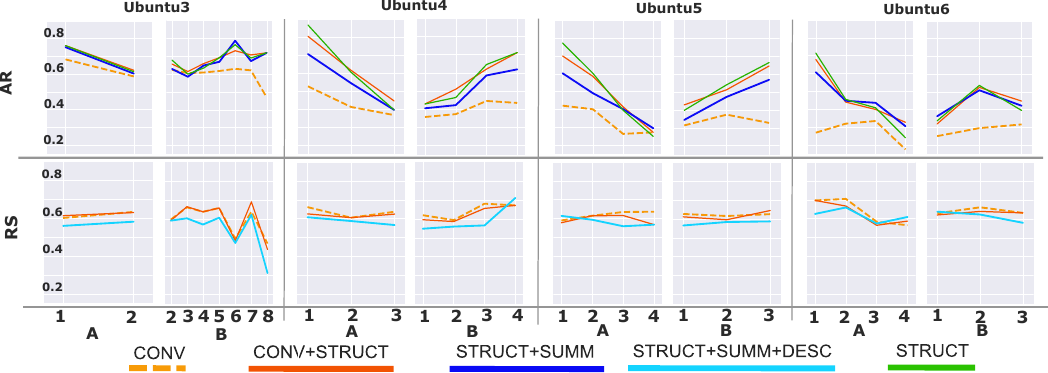}
    \caption{AR and RS accuracy results ($y$ axis) for the different values of $deg(u)$ (A) and $w_{avg}^o(u)$ (B) of the speaker node $u$ ($x$ axis). We report the performance of the three best input combinations for each task, plus \textsc{CONV} in AR which serves as text-only baseline. $w_{avg}^o(u)$ is rounded at the closest integer number.}
    \label{fig:structural_results}
\end{figure*}

\vspace{0.2cm}
\textbf{Structural Evaluation.}
In Figure \ref{fig:structural_results}, we present how the best run for each input combination varies in relation to the two network metrics introduced in Section \ref{sec:struc}, i.e. degree centrality $deg(u)$ and average outgoing weight $w_{avg}^o(u)$, where $u$ is the speaker node  (across all $4$ diagnostic datasets). More informally, $deg(u)$ (A in Figure \ref{fig:structural_results}, bottom) represents the number of users the next speaker has interacted with. Instead, $w_{avg}^o(u)$ (B in Figure \ref{fig:structural_results}, bottom) indicates the average number of messages sent to the users with whom the next speaker has interacted (in our graphs, we rounded it at the closest integer number).

In the AR task, combinations containing the interaction transcript (\textsc{+STRUCT}) exhibit similar patterns, while the \textsc{CONV} combination displays distinct trends compared to the other combinations. Notably, $deg(u)$ shows the strongest correlation with accuracy scores across all datasets: higher $deg(u)$ values consistently correspond to lower accuracy. Furthermore, the gap between the top-performing models (\textsc{CONV+STRUCT} and \textsc{STRUCT}) and others widens significantly at lower $deg(u)$ values. For example, while the \textsc{STRUCT} combination consistently ranks among the best in terms of macro-results, it is outperformed by (or comparable with) other combinations across all diagnostic datasets as $deg(u)$ increases. This shows that using in the prompt  \textsc{STRUCT}-only information is highly effective when the next speaker has spoken with few users in the transcripts (one or two), but it performs like the other combinations when the next speaker spoke with more than two users. As regards  $w_{avg}^o(u)$, the correlation with accuracy is less pronounced, but generally, higher $w_{avg}^o(u)$ values correspond to higher accuracy in models that use interaction transcripts as input (with some minor  fluctuations). 

In the RS task, we do not notice any clear correlation between $deg(u)$ and any increasing/decreasing behavior in the accuracies. Also for what concerns the gap among the models, there is no consistent trend across the different datasets. The same holds for  $w_{avg}^o(u)$. This suggests that the performances on the RS task are not related to the structural dimension.

\section{Discussion}
Our comparative evaluation shows three main findings.

\textbf{Input combination performance (RQ1).} Regarding the best-performing combinations, in AR, \textsc{STRUCT} and \textsc{CONV+STRUCT} consistently emerge as the top performers, with comparable results. This suggests that having only the interaction transcript is sufficient in our experimental setting for this task. Similarly, in RS, \textsc{CONV} and \textsc{CONV+STRUCT} consistently outperform other combinations, with the former widening the gap from the latter as more users are added. This indicates that having only the conversation transcript is adequate for the task in our experimental setup mostly based on textual information.
The inclusion of summary and/or the user description leads to a decline in performance. 
This may depend on the fact that \texttt{Llama2-13b-chat} sometimes is prone to generate bad summaries and descriptions, struggling to accurately capture the content of the conversation. On the other hand, it may also depend on the model difficulties to employ this information for classification. In the future, we plan to 
investigate this aspect with further analyses.

Interestingly, in both tasks, the gap between the best input combination and the best among the ones including summaries decreases as more users are involved (except for Ubuntu3 in the AR task): this suggests that summaries may be effective when dealing with a large number of users, as possible noise  introduced by the summary is equally challenging to dealing with complex conversations. Overall, user descriptions appear to be ineffective.

\textbf{Prompt Verbosity (RQ2).} 
Addressee recognition (AR), which benefits from structural information, shows greater sensitivity to prompts compared to response selection (RS), which is mostly a text-based task. 
This difference could be due to the similarity of RS to tasks used to pretrain LLMs. Indeed, RS is similar to a ``response generation'' task, where the perplexity of the two candidates is evaluated. 

As regards classification performance obtained with the different prompt versions, the verbose version of the prompt tends to be the best option for AR, probably because it helps the model in better capturing the structural information, which is crucial for this task. For RS, instead, there is no consistent improvement in using a more verbose prompt, probably because all the linguistic information necessary to perform the task is already expressed in the conversation. 

\textbf{Structural Complexity (RQ3)} 
Our structural analysis (Figure \ref{fig:structural_results}) reveals the limitations of relying solely on macro results, especially in the \textbf{AR task}. 

In AR, if we consider the correlation between classification accuracy and $deg(u)$, especially in Ubuntu4/5/6, we observe that the performance gaps between the best input combination (i.e., \textsc{STRUCT} for Ubuntu3/4, \textsc{CONV+STRUCT} for Ubuntu5/6) and \textsc{STRUCT+SUMM} combination in macro results is mainly driven by instances where the next speaker node interacts with only a few other users, for all diagnostic datasets. As the degree centrality increases, all combinations experience a general drop in performance. 
This analysis underscores that macro results offer only a surface-level understanding of the model's capabilities and are heavily influenced by dataset characteristics. A closer examination reveals that the best input combinations perform very well in simpler conversations, with limited generalization to more complex interaction structures. 
A model being able to effectively capture both structural and linguistic information should ideally show less performance degradation at increasing degree centrality, rather than performing well only on samples with lower degree centrality.
Regarding $w_{avg}^o(u)$, it suggests that having more messages directed towards the involved users may help determine the last addressee. However, as shown by the performance of \textsc{STRUCT}, this is not due to message information. Nevertheless, this could still be an effect of the degree centrality, as the conversation length is fixed at 15 messages and higher values of $w_{avg}^o(u)$ likely correspond to lower values of $deg(u)$. 

In AR, if we consider only conversations with a complex structure, we observe that the inputs that address data minimization (e.g. those using conversation summary) reach a performance close to the best performing input combination. Instead, the worse performance with data minimization input is obtained when classifying examples with low structural complexity. Therefore, in the future it may be worth focusing on simple structures and try to address this performance gap, understanding its causes.

The structural analysis of the \textbf{RS results} indicates that performance for this task is largely unaffected by network metrics. This observation can be interpreted in two ways: firstly, information on the structure of the conversation may not be relevant when selecting a response. Alternatively, the model might effectively infer the conversation flow based solely on message content, maintaining consistent performance regardless of the ``node complexity''.

For both tasks (AR and RS), we analysed other node metrics (i.e., closeness centrality and clustering coefficient). Such metrics showed high correlation with the degree centrality, and for this reason we do not report them here.

\section{Conclusions}
In this study, we evaluate the zero-shot performance of an LLM (i.e., \texttt{Llama2-13b-chat}) on two tasks based on multi party conversations, namely response selection and addressee recognition. Our goal is to provide an in-depth analysis of different experimental settings tested for the two tasks, which include three different prompt types and  
six configurations to model the conversation text and its structure. 
Our analysis is performed on four diagnostic datasets with a fixed number of users. For each of them, we compute  two network metrics, i.e. degree centrality and average outgoing weight, to analyse how structural complexity interacts with classification performance. We devote particular attention to evaluating how strategies to replace the original conversation text could be effectively used in the prompts. This is very relevant to ensure a safe use of MPC corpora: if the same classification performance could be achieved by removing the original conversation, data resharing would not imply the risk of making personal or sensitive data available. Furthermore, malicious use of MPCs, for example using them to train models with fake personas, would not be possible. Although promising, this research direction has not achieved fully satisfactory results.

The goal of our work is not much to yield  the best possible classification accuracy on AR and RS, but rather to provide an in-depth analysis of the possible dimensions contributing to the classifier performance on the two tasks. We believe that the interplay between textual and structural information in MPCs should be better analysed in current evaluations, merging contributions from NLP and the network science community.  

\section{Limitations}
The findings presented in this work are based only on subsets of a single dataset, the Ubuntu Internet Relay Chat corpus. This choice is due to the fact that many multi-party datasets lack sufficient variety in terms of structure and addressee labels, which are necessary for performing in-depth diagnostic analyses. Additionally, all the conversations in our dataset have the same length ($15$ turns), allowing us to exclude conversation length as a variable. During the development of this work, we took other datasets into account as possible candidates for our experiments, but when we analysed them more in depth we found that they presented neither the structural characteristics which are necessary to build diagnostic datasets, nor the necessary amount of data to perform a good diagnostic analysis.
However, for future research, it would be interesting to introduce other types of diagnostic datasets, for example extracted from different social media or dealing with a diverse set of topics.
Moreover, our experiments were conducted using only one instruction-based LLM in a zero-shot setting, as our primary goal was to present a novel evaluation pipeline. Furthermore, we evaluated classification performance based on the best run for each model and combination. However, it is important to note that claiming general capabilities of the model based on these results would be scientifically inaccurate. Comparing different LLMs would be necessary to better prove the generalisation of our approach. 

\section{Ethics Statement}
In conducting this research, we prioritized the privacy and ethical management of data. Although the original dataset (UbuntuIRC) is freely distributed online and includes the original usernames, we chose to anonymize the data by replacing each username with a name from a set of ungendered names (details are provided in the Appendix \ref{sec:app_prompt_design}). This approach helps to protect the identity of the users and reduces potential biases associated with gendered names. Furthermore, we explored two alternative representations of the conversation transcripts to investigate the feasibility of working with different elements compared to the original messages. This is in line with current policies that encourage researchers to find methods that enhance user privacy and minimize biases. Future research could benefit from using summaries and/or user descriptions, which would allow the distribution of textual data while making it nearly impossible to train chatbots that could imitate specific users.
Regarding reproducibility, our experiments were conducted in a zero-shot setting without fine-tuning. This ensures that our methodology can be replicated efficiently. Specifically, our experiments can be reproduced within a few hours on a single GPU with 48GB of VRAM and a batch size of one, using a model that is available online.

\section*{Acknowledgments}
The work of BL was partially supported by the NextGenerationEU Horizon Europe Programme, grant number 101120237 - ELIAS and grant number 101120763 - TANGO. 
BL and ST were also supported by the PNRR project FAIR - Future AI Research (PE00000013). NP’s activities are part of the network of excellence of the European Laboratory for Learning and Intelligent Systems (ELLIS).

\bibliography{anthology,custom}

\newpage

\appendix

\begin{figure}[ht]
    \centering
    \includegraphics[width=0.45\textwidth]{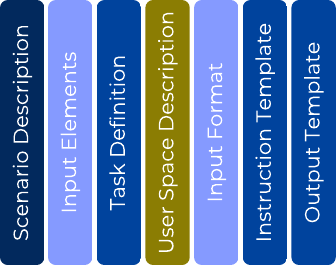}
    \caption{Graphical representation of the system prompt organization.}
    \label{fig:system_prompt}
\end{figure}

\section{Prompt schemes and combinations} \label{sec:app_prompt_design}

In our experimental setup, we establish a fixed template for all system prompts, as shown in Figure \ref{fig:system_prompt}, consisting in $7$ sections:

\begin{itemize}
    \item \textbf{Scenario Description}: describes the scenario, defining messages and interactions between speakers and addressees;

    \item \textbf{Input Elements}: lists the input elements provided to the model according to the input combination, for example CONV, CONV+STRUCT, STRUCT+SUMM, etc.;

    \item \textbf{Task Definition}: defines the task to be performed (response selection, addressee recognition, generating summaries, or generating user descriptions);

    \item \textbf{User Space Description}: defines which users are involved as speakers or addressees;

    \item \textbf{Input Format}: specifies how the input elements are presented in the prompt;

    \item \textbf{Instruction Template}: details how the task instruction command is written in the prompt;

    \item \textbf{Output Template}: defines how the generated output should be organized. 
    
\end{itemize}

\begin{figure}
    \centering
    \includegraphics[width=0.45\textwidth]{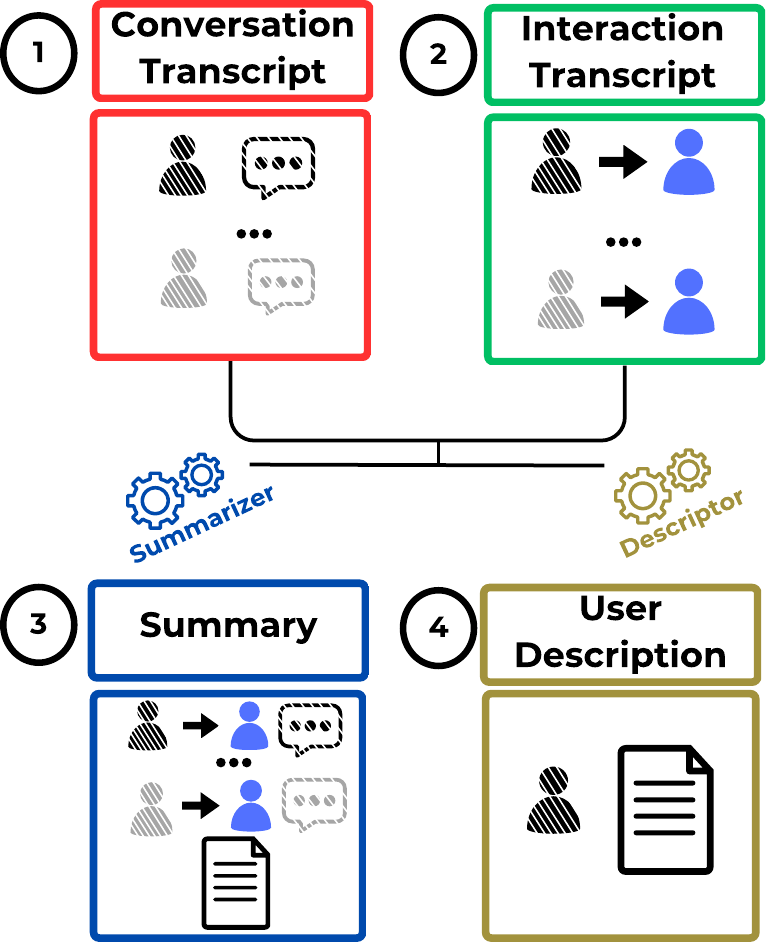}
    \caption{\textbf{Experimental setup}. First we create the conversation transcript (1) and the interaction transcript (2). From these, we extract the summary and the user description by using a specifically prompted LLM (3,4).}
    \label{fig:input_info}
\end{figure}

The Scenario Description and User Space Description remain consistent across all tasks and combinations; three sections, i.e. Task Definition, Instruction Template and Output Template, vary depending on the specific task (e.g., AR, RS, summarisation, description); two sections, i.e. Input Elements and Input Format, are constructed modularly based on the chosen input combination (e.g., CONV, STRUCT, SUMM, DESC). In Figure \ref{fig:input_info} we report how the different pieces of input information are related to each other.

There is an ongoing discussion about evaluating instruction-based models particularly considering the high sensitivity of their performance to different levels of prompt verbosity. 
For this reason, we identify a first dimension across all task, calling it ``prompt scheme''. Each prompt scheme consists in totally writing all the sections from scratch and recreating the prompts across tasks and combinations. 

\begin{figure*}[ht]
    \centering
    \includegraphics[width=0.85\textwidth]{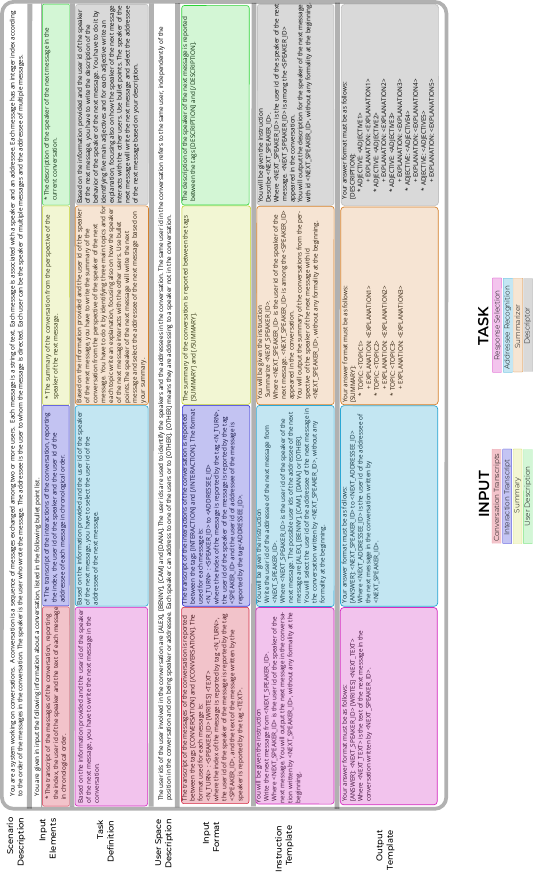}
    \caption{First prompt scheme (verbose version).}
    \label{fig:prompt_scheme0}
\end{figure*}

\begin{figure*}[ht]
    \centering
    \includegraphics[width=0.8\textwidth]{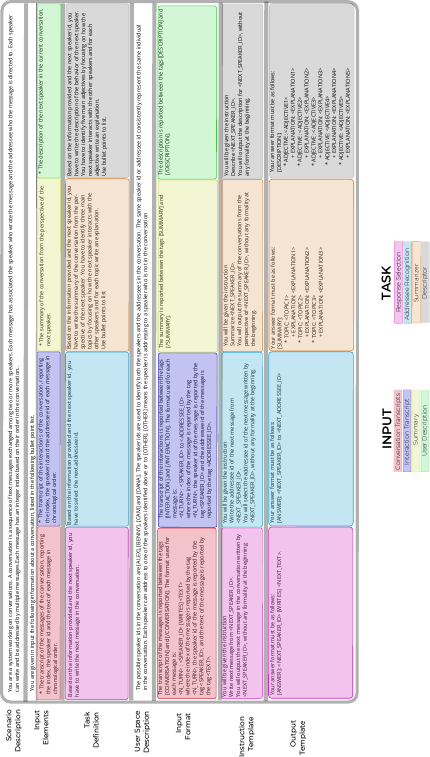}
    \caption{Second prompt scheme (medium version).}
    \label{fig:prompt_scheme1}
\end{figure*}

\begin{figure*}[ht]
    \centering
    \includegraphics[width=0.78\textwidth]{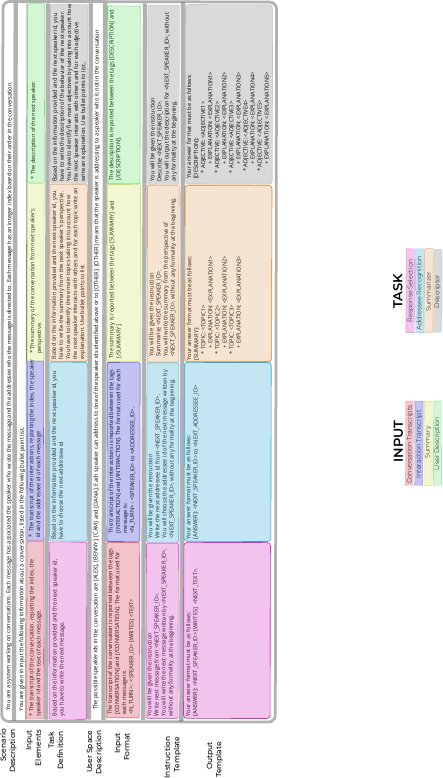}
    \caption{Third prompt scheme (concise version).}
    \label{fig:prompt_scheme2}
\end{figure*}

We create three prompt schemes corresponding to three different levels of prompt verbosity. The first, reported in Figure \ref{fig:prompt_scheme0} is extremely precise and detailed, the second (Figure \ref{fig:prompt_scheme1}) gives some concepts for granted and the third (Figure \ref{fig:prompt_scheme2}) is the most implicit. One clear example is in the Instruction Template for the AR task. It evolves from \textit{``Write the user id of the addressee of the next message''}, to \textit{``Write the addressee id of the next message''}, and finally to \textit{``Write the next addressee id''}. 

\begin{figure*}[ht]
    \centering
    \includegraphics[width=0.75\textwidth]{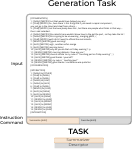}
    \caption{Input information and instruction command for generation.}
    \label{fig:input_prompt_gen}
\end{figure*}

\begin{figure*}[ht]
    \centering
    \includegraphics[width=0.75\textwidth]{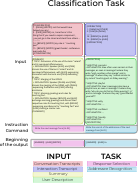}
    \caption{Input information and instruction command for classification. Here we also add a ``beginning of input''section.}
    \label{fig:input_prompt_class}
\end{figure*}

\begin{figure*}[ht]
    \centering
    \includegraphics[width=0.72\textwidth]{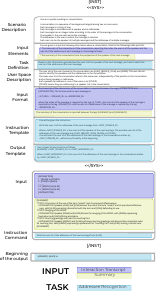}
    \caption{Example of final prompt for the \textsc{STRUCT+SUMM} combination in the AR task, following the first prompt scheme.}
    \label{fig:prompt_example}
\end{figure*}

After creating the system prompt, we concatenate the input information and the instruction command, as shown in Figure \ref{fig:input_prompt_gen} for generation tasks and in Figure \ref{fig:input_prompt_class} for classification tasks, to form the final prompt. An example of this process for the \textsc{STRUCT+SUMM} combination in the AR task is shown in Figure \ref{fig:prompt_example}. 

\begin{figure*}[ht]
    \centering
    \includegraphics[width=0.75\textwidth]{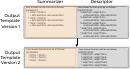}
    \caption{Versions of output template tested for generating summary/user description.}
    \label{fig:output_template}
\end{figure*}

We identify the second dimension only for the generation of summary/user description task. Once we fix the prompt scheme, we test two output templates, as shown in Figure \ref{fig:output_template}, while keeping all the other sections fixed. Since the results are similar (details in Section \ref{sec:app_output_template}) we 
mention only the first version for the arguments in the main body of the paper.

For user anonymization, we replace each original username with one of the following ungendered user tags:
[ALEX], [BENNY], [CAM], [DANA], [ELI], and [FREDDIE]. The tag [ALEX] is always assigned to the next speaker. 

\section{Task details and formalization}\label{sec:appendix_task}

\begin{table*}
\centering
\small
\begin{tabular}{|p{3.2cm}|c|c|c|c|c|}
\hline
\textbf{COMBINATION} & \textbf{PROMPT SCHEME} & \textbf{UBUNTU3} & \textbf{UBUNTU4} & \textbf{UBUNTU5} & \textbf{UBUNTU6} \\
\hline
CONV & verbose & 0.613 & 0.414 & 0.352 & 0.277 \\
     & medium  & 0.582 & 0.409 & 0.344 & 0.283 \\
     & concise  & 0.595 & 0.416 & 0.298 & 0.289 \\
\hline
CONV+STRUCT & verbose & 0.660 & 0.584 & 0.525 & 0.449 \\
            & medium  & 0.609 & 0.501 & 0.513 & 0.431 \\
            & concise  & 0.571 & 0.477 & 0.465 & 0.400 \\
\hline
STRUCT+SUMM & verbose & 0.623 & 0.517 & 0.448 & 0.397 \\
              & medium  & 0.644 & 0.491 & 0.465 & 0.429 \\
              & concise  & 0.617 & 0.441 & 0.433 & 0.374 \\
\hline
STRUCT+DESC & verbose & 0.637 & 0.499 & 0.456 & 0.406 \\
              & medium  & 0.604 & 0.457 & 0.442 & 0.380 \\
              & concise  & 0.618 & 0.458 & 0.404 & 0.323 \\
\hline
STRUCT+SUMM+DESC & verbose & 0.628 & 0.472 & 0.429 & 0.383 \\
                     & medium  & 0.620 & 0.455 & 0.444 & 0.374 \\
                     & concise  & 0.607 & 0.433 & 0.417 & 0.323 \\
\hline
STRUCT & verbose & 0.654 & 0.572 & 0.537 & 0.454 \\
       & medium  & 0.626 & 0.515 & 0.498 & 0.434 \\
       & concise  & 0.573 & 0.487 & 0.438 & 0.389 \\
\hline
\end{tabular}
\caption{Table of Accuracies in addressee recognition across prompt schemes and input combinations}
\label{tab:ubuntu_scores_ar_prompt_schemes}
\end{table*}

\begin{table*}
\centering
\small
\begin{tabular}{|p{3.2cm}|c|c|c|c|c|}
\hline
\textbf{COMBINATION} & \textbf{PROMPT SCHEME} & \textbf{UBUNTU3} & \textbf{UBUNTU4} & \textbf{UBUNTU5} & \textbf{UBUNTU6} \\
\hline
CONV & verbose & 0.625 & 0.627 & 0.619 & 0.640 \\
     & medium  & 0.624 & 0.619 & 0.613 & 0.646 \\
     & concise  & 0.612 & 0.617 & 0.610 & 0.649 \\
\hline
CONV+STRUCT & verbose & 0.626 & 0.611 & 0.602 & 0.620 \\
            & medium  & 0.626 & 0.609 & 0.606 & 0.631 \\
            & concise  & 0.618 & 0.616 & 0.590 & 0.629 \\
\hline
STRUCT+SUMM & verbose & 0.572 & 0.570 & 0.569 & 0.626 \\
              & medium  & 0.575 & 0.556 & 0.573 & 0.614 \\
              & concise  & 0.564 & 0.572 & 0.587 & 0.623 \\
\hline
STRUCT+DESC & verbose & 0.565 & 0.553 & 0.540 & 0.597 \\
              & medium  & 0.553 & 0.565 & 0.550 & 0.586 \\
              & concise  & 0.542 & 0.562 & 0.538 & 0.606 \\
\hline
STRUCT+SUMM+DESC & verbose & 0.576 & 0.570 & 0.573 & 0.623 \\
                     & medium  & 0.574 & 0.570 & 0.575 & 0.614 \\
                     & concise  & 0.573 & 0.583 & 0.585 & 0.620 \\
\hline
\end{tabular}
\caption{Table of Accuracies in response selection across prompt schemes and input combinations}
\label{tab:ubuntu_scores_rs_prompt_schemes}
\end{table*}

\subsection{Formalization of an MPC}
Given a conversation $C = (M, U)$, $M = \{m_1, m_2, ..., m_n\}$ is the set of chronologically ordered messages (message $m_i$ appeared before $m_j$ in the conversation if $i < j$) and $U = \{u_1, u_2, ..., u_p\}$ is the set of users occurred in $C$. Each message $m_i$ is assigned a ordered pair $(u_j, u_k)$ s.t. $u_j$ is the speaker of $m_i$ and $u_k$ is the addressee of $m_i$, so $u_j = S(m_i)$ and $u_k = A(m_i)$.

\subsection{Classification using CPPL}

Given the task $T \in \{RS, AR\}$, a classification prompt $p_T$, and the set of candidate responses $R_T = \{r_1, ..., r_m\}$, we extract as output the candidate with minimum conditional perplexity $\min CPPL(r_i|p), i \in [1,m]$, where

\begin{equation*}
    CPPL(r_i|p) = \frac{1}{P(r_i|p)^{1/|r_i|}}
\end{equation*}

according to the probability distribution of the model.

From the output CPPL, we can obtain a probability distribution over the set of candidates, so 

\begin{equation*}
    P(r_k) = \frac{1/CPPL(r_k)}{\sum_{r_i\in R_T}1/CPPL(r_i)}
\end{equation*}

Analysing the correlation between the probability of the output target and the network metrics leads to the same conclusions obtained considering accuracy values.

\subsection{Results in detail}

In this section we report all the results for each prompt scheme and combination, on both AR (Table \ref{tab:ubuntu_scores_ar_prompt_schemes}) and RS (Table \ref{tab:ubuntu_scores_rs_prompt_schemes}). It is the tabular version of the graphs in Figure \ref{fig:macro_results}, where only the best run is reported, and the most detailed version of Table \ref{tab:prompt_gap}, where we report the relative gap between the best run and the average among the $3$ prompt schemes (from each cell, we extract only the gap).

\subsection{Effect of different output templates in generating summary/user description prompt}\label{sec:app_output_template}

\begin{figure*}[ht]
    \centering
    \includegraphics[width=0.98\textwidth]{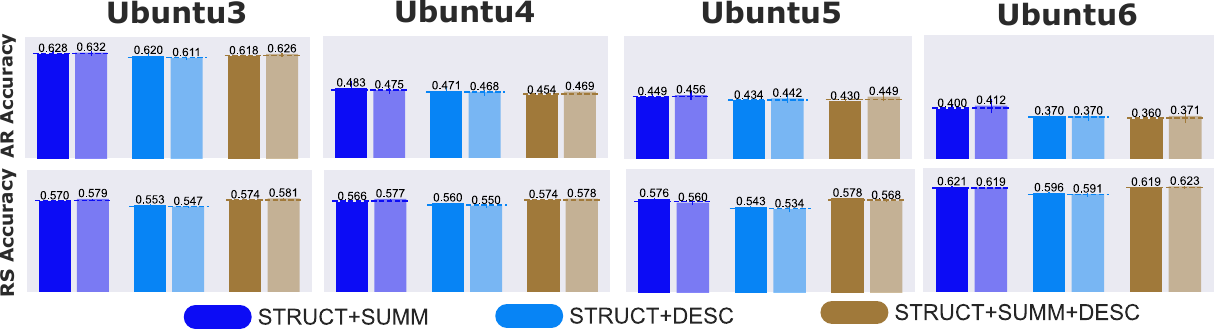}
    \caption{Results across the two generation of summary/user description output template.}
    \label{fig:graph_output_template}
\end{figure*}

In Figure \ref{fig:graph_output_template} we report the results across diagnostic datasets and tasks for \textsc{STRUCT+SUMM}, \textsc{STRUCT+DESC}, and \textsc{STRUCT+SUMM+DESC} by averaging the results across the different prompt schemes but with the same output template. The average between the two different output templates does not differ much, with a maximum difference of $1.9\%$ in the AR task-Ubuntu5 for \textsc{STRUCT+SUMM+DESC}, between the averages with same combination but different output templates. Nevertheless, we examine all combinations by averaging the results across the output template, obtaining results consistent with the ones presented in Section \ref{sec:results}. We focus therefore on one output template for the sake of simplicity.

\subsection{Technical details}
For our experiments we use a single A40 GPU with 48GB Memory. With such GPU, it is possible to use \texttt{Llama2-13b-chat} only in inference with a batch size of $1$. We used the \texttt{Llama-2-13b-chat-hf} version provided by HuggingFace\footnote{\url{https://huggingface.co/meta-llama/Llama-2-13b-chat-hf}}. We use Copilot\footnote{\url{https://github.com/features/copilot}} as a coding assistant and ChatGPT \cite{openai_chatgpt} as a writing assistant only to improve the style of the text.

\end{document}